\documentclass[lettersize,journal]{IEEEtran}
\usepackage{amsmath,amsfonts}
\usepackage{algorithmic}
\usepackage{algorithm}
\usepackage{array}
\usepackage[caption=false,font=normalsize,labelfont=sf,textfont=sf]{subfig}
\usepackage{textcomp}
\usepackage{stfloats}
\usepackage{url}
\usepackage{verbatim}
\usepackage{graphicx}
\usepackage{cite}
\usepackage{algorithm}
\usepackage{algorithmic}
\usepackage{amssymb}
\usepackage{multirow}
\usepackage{amsmath} 

\pdfoutput=1

\begin{document}

\title{FLEX-CLIP: \textbf{F}eature-\textbf{L}evel G\textbf{E}neration Network Enhanced CLIP for \textbf{X}-shot Cross-modal Retrieval}

\author{Jingyou Xie$^{\dag}$, Jiayi Kuang$^{\dag}$, Zhenzhou Lin, Jiarui Ouyang, Zishuo Zhao, and Ying Shen$^{\ast}$ 
\thanks{$^{\dag}$ indicates equal contribution, * indicates corresponding author}
\thanks{Jingyou Xie, (e-mail: Xiejy73@mail2.sysu.edu.cn). Jiayi Kuang, (e-mail: kuangjy6@mail2.sysu.edu.cn). Zhenzhou Lin, (e-mail: linzhzh6@mail2.sysu.edu.cn). Jiarui Ouyang, (e-mail: ouyjr@mail2.sysu.edu.cn). Zishuo Zhao, (e-mail: zhaozsh@mail2.sysu.edu.cn). and Ying Shen, (e-mail:sheny76@mail.sysu.edu.cn) are with the School of Intelligent Systems Engineering, Sun Yat-sen University, Shenzhen, 518107 China.}}

\markboth{Journal of \LaTeX\ Class Files,~Vol.~14, No.~8, August~2021}%
{Shell \MakeLowercase{\textit{et al.}}: A Sample Article Using IEEEtran.cls for IEEE Journals}

\IEEEpubid{0000--0000/00\$00.00~\copyright~2021 IEEE}

\maketitle

\begin{abstract}
Given a query from one modality, few-shot cross-modal retrieval (CMR) retrieves semantically similar instances in another modality with the target domain including classes that are disjoint from the source domain. Compared with classical few-shot CMR methods, vision-language pretraining methods like CLIP have shown great few-shot or zero-shot learning performance. However, they still suffer challenges due to (1) the feature degradation encountered in the target domain and (2) the extreme data imbalance. To tackle these issues, we propose FLEX-CLIP, a novel Feature-level Generation Network Enhanced CLIP. FLEX-CLIP includes two training stages. In multimodal feature generation, we propose a composite multimodal VAE-GAN network to capture real feature distribution patterns and generate pseudo samples based on CLIP features, addressing data imbalance. For common space projection, we develop a gate residual network to fuse CLIP features with projected features, reducing feature degradation in X-shot scenarios. Experimental results on four benchmark datasets show a 7\%-15\% improvement over state-of-the-art methods, with ablation studies demonstrating enhancement of CLIP features.
\end{abstract}


\section{Introduction}

The rise of massive multimodal data, encompassing text and images \cite{baltruvsaitis2018multimodal,bayoudh2022survey, xu2023multimodal}, has led to increasing demand for Cross-Modal Retrieval (CMR), which enables searching for similar instances in one modality using query data from another modality \cite{zhen2019deep,wei2016cross,rasiwasia2010, wang2017adversarial,chun2021probabilistic}. Unlike single-modal retrieval, cross-modal retrieval faces the challenge of a “modality gap”, stemming from inconsistent representations and differing distributions between modalities \cite{wang2015joint, zhen2019deep,wei2016cross,xu2019deep,gao2020survey}.

To bridge this gap, current methods typically map samples from different modalities into a common space \cite{wei2016cross,xu2019deep,deng2018triplet,wang2019matching,wang2017adversarial,xie2022token}. The training process learns this common space from the training instances, allowing direct calculation of cross-modal similarities for the test data \cite{chun2021probabilistic,zhang2021hcmsl,bogolin2022cross}. However, these methods require the training and test data to share the same set of classes, limiting their extensibility and generalization to new categories \cite{xu2020correlated,kaur2021comparative}.

\begin{figure}
    \centering
    \includegraphics[width=0.9\linewidth]{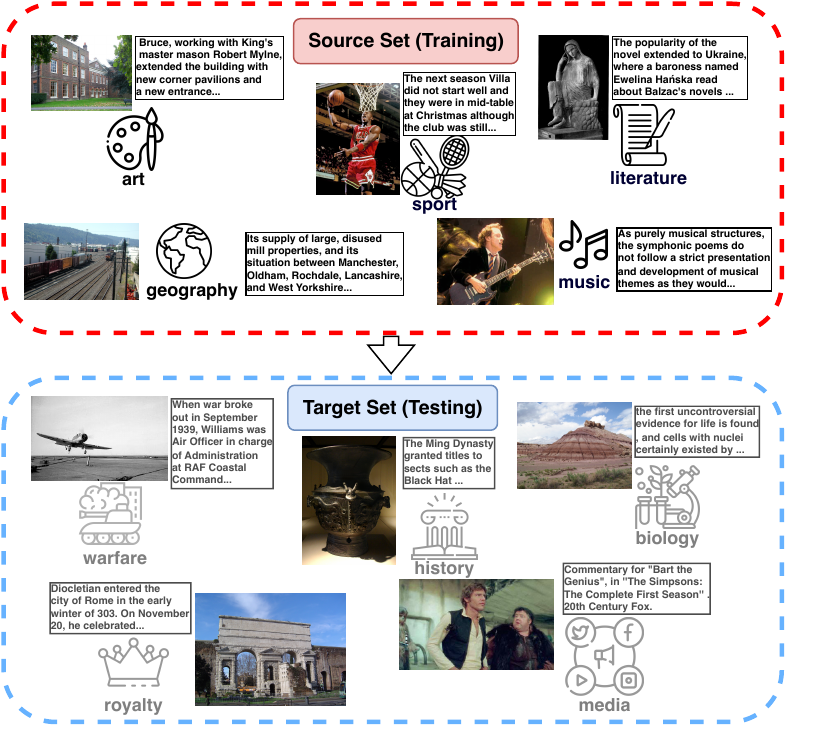}
    \caption{The illustration of the zero-shot cross-modal retrieval dataset setting.}
    \label{zero}
\end{figure}

These approaches also struggle in few-shot scenarios, particularly due to the challenge of extreme data imbalance when the retrieval classes in the source domain have little or no overlap with those in the target domain \cite{lin2020learning, pourpanah2022review}, as shown in Figure \ref{zero}. To address the new classes in the target domain, many methods draw inspiration from few-shot learning, which leverages word embeddings of class label information, as depicted in Figure \ref{fig1} (a) \cite{socher2013zero,frome2013devise,akata2013label,chi2018dual,sun2021research}. These methods seek to establish a shared latent space between input multimodal features and class label embeddings, or they generate pseudo-samples based on target domain class embeddings to mitigate data imbalance \cite{chi2019zero, xu2020correlated,liu2018generalized,chou2020adaptive}. Given the strong few-shot learning performance of vision-language pretraining models (VLP) on various multimodal tasks \cite{hafner2021clip,Rao2023VL}, some approaches incorporate CLIP \cite{radford2021learning} as the feature extractor to enhance cross-modal retrieval.

Despite advancements in few-shot and zero-shot cross-modal retrieval, significant challenges remain:

\begin{figure*}[t]
    \centering
    \includegraphics[width=0.9\linewidth]{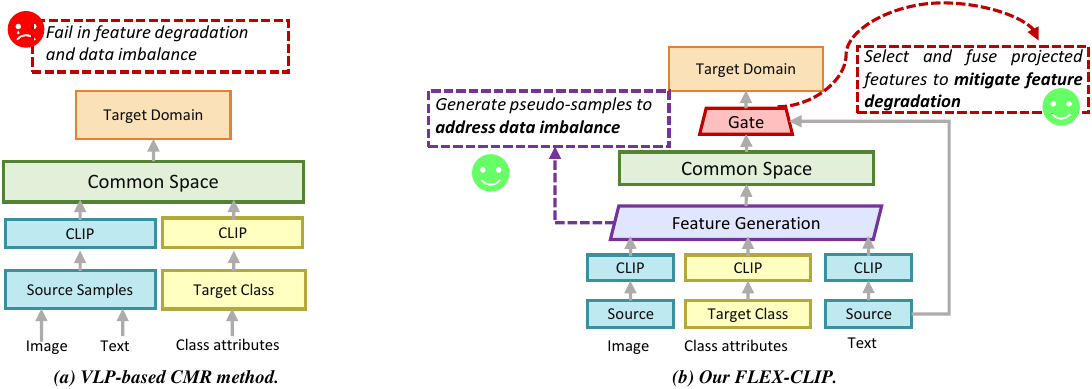}
    \caption{An illustration of classical zero-shot CMR methods VLP-based method, and our method.}
    \label{fig1}
\end{figure*}

1)\textbf{\textit{Feature Degradation in the Target Domain:}} While CLIP-based CMR methods exhibit remarkable few-shot learning performance, the common space mapping of features extracted by CLIP in few-shot settings often leads to feature degradation. This issue arises when the model struggles to effectively transfer the learned common space mapping from the source domain to the target domain, resulting in the mapped features performing worse than the original CLIP-extracted features.

2)\textbf{\textit{Extreme Data Imbalance}}: The absence of target domain instances during training makes it challenging to model the cross-modal correlations in the target set \cite{han2021contrastive,Ma2022,Xian2019}. Some methods attempt to generate pseudo-samples based on target domain class embeddings to mitigate data imbalance. However, due to the disparity in the number of samples between the source and target domains, the model often fails to accurately capture the true feature distribution of the target domain. This results in a bias toward the source domain during training and suboptimal performance on the target domain \cite{chen2021free,xu2020attribute}.

To address these challenges, we propose FLEX-CLIP, a novel \textbf{F}eature-\textbf{L}evel G\textbf{E}neration Network enhanced CLIP for \textbf{X}-shot Cross-modal Retrieval. FLEX-CLIP comprises two key stages: (1) multimodal feature generation, and (2) common space projection. In the feature generation stage, inspired by classical zero-shot CMR methods, we design a composite cross-modal generation architecture. This architecture leverages the strengths of both VAE and GAN networks to tackle data imbalance and enhance CLIP features. The GAN generates pseudo-samples based on class embeddings, while the VAE captures feature distribution patterns by encoding and reconstructing real samples. In the common space projection stage, real samples from the source domain and generated samples from the target domain are fed into two projectors to obtain common features. To effectively utilize the semantic information in CLIP, we design a gated residual network that selectively fuses original features with the mapped features, thereby significantly addressing the feature degradation problem.

Notably, unlike previous approaches that integrate data generation and common space mapping into a single framework, we train these two processes separately to ensure stable results. Our main contributions are as follows:

\begin{itemize}
\item We introduce a novel \textbf{F}eature-\textbf{L}evel G\textbf{E}neration Network to enhance CLIP features for \textbf{X}-shot CMR, effectively addressing the extreme data imbalance problem in few-shot learning within CLIP.

\item We design a gated network that adaptively selects and fuses projected features with original CLIP features, fully leveraging semantic knowledge and mitigating feature degradation.

\item We conduct extensive experiments on four widely-used datasets in X-shot scenarios (0, 1, 3, 5, 7-shot) and demonstrate that FLEX-CLIP significantly outperforms strong baselines in few-shot cross-modal retrieval tasks by up to 7.9\%.

\end{itemize}

\section{Related Work}

\subsection{Cross-modal Retrieval}

Cross-modal retrieval seeks to identify semantically similar instances in one modality using a query from another, with image-text and image-sketch retrieval being two common scenarios \cite{zhen2019deep,semedo2019diach,wei2016cross}. Traditional approaches achieve this by jointly modeling the text and image components of multimedia documents, linearly projecting features from different modalities into a shared space \cite{rasiwasia2010,Hotelling1992}. Deep neural networks (DNNs) have further advanced this field by reducing the gap between modalities and identifying maximally correlated embedding spaces \cite{wang2015joint,xu2019deep,chen2022cross}. While DNN-based methods can integrate various features into a common space, they sometimes struggle to model heterogeneous data distributions with inconsistent semantics. Recent research has leveraged Generative Adversarial Networks (GANs) to address these challenges, using cross-modal generators to reduce semantic discrepancies and modality discriminators to distinguish features across modalities \cite{deng2018triplet,wang2019matching,wang2017adversarial}. Additionally, vision-language pretraining (VLP) models like CLIP \cite{radford2021learning} have shown significant promise in cross-modal retrieval, though they may suffer from feature degradation when encountering new classes in the target domain.

\begin{figure*}
    \centering
    \includegraphics[width=1 \linewidth]{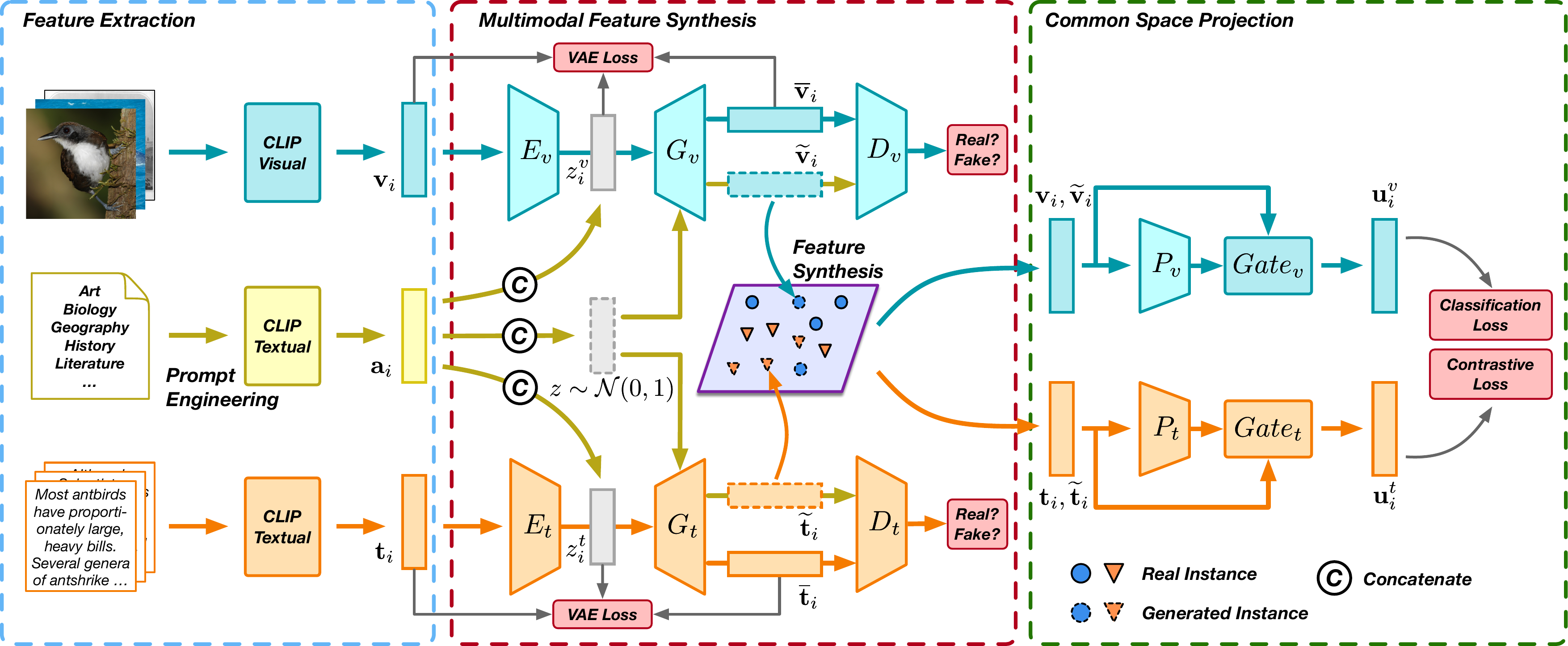}
    \caption{The whole architecture of our proposed model FLEX-CLIP. $\mathbf{v_i}$ and $\mathbf{t_i}$ are a pair of image and textual features, respectively. $\mathbf{y_i}$ represents the category label, and $\mathbf{a_i}=attri(y_i)$ is the corresponding category attribute feature of this sample. $E_* (\cdot)$, $G_* (\cdot)$, $D_* (\cdot)$ denote encoder, generator, discriminator respectively, and $*=\{v,t\}$ denotes the image and text modalities.}
    \label{fig2}
\end{figure*}

\subsection{Zero-shot Cross-modal Retrieval}

Although traditional methods perform well, they often struggle with zero-shot retrieval due to the absence of unseen target classes during training \cite{xu2020correlated}. To address this, various studies have explored class embeddings to enable zero-shot learning \cite{lin2020learning,frome2013devise,akata2013label,chi2018dual}. For instance, Chi et al. introduced DADN to address heterogeneous distributions and inconsistent semantics between seen and unseen classes, leveraging word embeddings to enhance knowledge transfer \cite{chi2019zero}. Recent work has increasingly focused on feature generation techniques to mitigate the modality gap and data imbalance issues \cite{han2021contrastive,Ma2022,Xian2019}, employing generative models like GANs and Variational Autoencoders (VAEs) to capture multimodal data structures, as seen in LCALE \cite{lin2020learning} and AAEGAN \cite{xu2021zero}. To handle inconsistent semantics between source and target sets, Xu et al. proposed TANSS, which includes two modality-specific semantic learning subnetworks and a self-supervised semantic learning subnetwork to generalize across datasets \cite{xu2020Ternary}. However, these approaches often overlook the impact of data reconstruction in generative models, leading to a misleading feature distribution. MDVAE \cite{tian2022multimodal} was introduced to address this by learning disentangled modality-invariant and modality-specific features through coupled VAEs. Nonetheless, generative models frequently lack scalability for retrieving cross-modal data from new classes due to feature degradation. JFSE \cite{xu2022joint} revisits adversarial learning in existing cross-modal GAN methods, employing coupled conditional Wasserstein GAN modules to synthesize meaningful and correlated multimodal features.


\section{Methodology}
\subsection{Problem Formulation}
In this work, we focus on zero-shot and few-shot cross-modal retrieval between text and images. Given an instance set of source domain $X_{s}=\{\textbf{x}_{i}\}_{i=1}^{n_{s}}$ and an instance set of target domain $X_{t}=\{\textbf{x}_{j}\}_{j=1}^{n_{t}}$, they contains $n_{s}$ and $n_{t}$ samples respectively. Taking the source domain set $X_s$ as an example, the $i$-th sample is defined as $\textbf{x}_{i}=(\textbf{v}_{i}, \textbf{t}_{i}, \textbf{a}_{i}, \textbf{y}_{i})$, where $\textbf{v}_{i}\in\mathbb{R}^{d_{v}}$ and $\textbf{t}_{i}\in\mathbb{R}^{d_{t}}$ are image-text pair that have the same semantic information. $\textbf{y}_{i}$ is the class label, and $\textbf{a}_{i}=attri(\textbf{y}_{i})$ is the corresponding class attribute feature of $\textbf{y}_{i}$. It is worth noting that the class label set in target domain $Y_{t}=\{\textbf{y}_{j}\}_{j=1}^{c_{t}}$ is completely different from the ones in source domain$Y_{s}=\{\textbf{y}_{i}\}_{i=1}^{c_{s}}$, which indicates that $Y_{t}\cap Y_{s}=\varnothing$.

In the few-shot scenario, the model can be trained using samples from both the source domain $X_s$ and a small number of samples from the target domain $X_t$, and then tested on the target domain $X_t$. In the zero-shot scenario, the model can only be trained by using the samples in the source domain $X_s$ and then tested on the target domain $X_t$. 

\subsection{Model Architecture}
The general framework of our model is shown in Figure \ref{fig2}. In Feature Extraction (Sec. \ref{Feature Extraction}), We utilize three pre-trained feature extractors to extract features from images, class labels, and text, respectively. In Multimodal Feature Generation (Sec. \ref{Multimodal Feature Synthesis}), we propose a composite VAE-GAN network. After the features are encoded via their respective encoders, the generator decodes the VAE reconstructed features respectively, and the VAE loss is applied to learn the distribution pattern of features. In addition, the generators generate pseudo-sample features based on category attributes, which are inputted into the discriminator along with the VAE reconstructed features. In Common Space Projection (Sec. \ref{Common Space Projection}), we first propose a projector that maps the real original features common space together, and then propose a residual gate network to selectively fuse the original features and the projected features. Finally, we design three loss functions to narrow the modality gap and facilitate the retrieval performance.

\subsection{Feature Extraction} \label{Feature Extraction}
Before the process of feature generation, we need to map the origin text and image data in the multimodal dataset from the real scenario spaces to their original feature spaces, respectively. We adopt the CLIP \cite{radford2021learning} model as the feature extraction framework. For image modalities, we leverage the CLIP image encoder with ResNet-50 \cite{he2016deep} backbone to extract 1,024-d features as the original image feature $\textbf{v}$. For text modality, we apply a CLIP text encoder with BERT \cite{kenton2019bert} backbone to extract 1,024-d features as the original text feature $\textbf{t}$. For class embedding, different from the most recent used Word2Vec as the extractor, we introduce prompt engineering to construct the category words as text statements and then utilize the CLIP text encoder for attribute embedding extraction to obtain the 1024-d features as category attribute embedding $\textbf{a}$, which ensures the consistency between image-text-category attributes.

\subsection{Multimodal Feature Generation}\label{Multimodal Feature Synthesis}
In contrast with methods such as image generation for pixel-level images or text generation for word-level text, we aim to generate feature-level multimodal data for downstream multimodal projector. For each modality, we respectively construct a composite VAE-GAN network to fully leverage the advantages of VAE in capturing sample feature distribution and GAN in stabilizing feature generation.  

Each network consists of three main components: an encoder $E_{*}(\cdot)$, a generator $G_{*}(\cdot)$, and a discriminator $D_{*}(\cdot)$, where $*=\{v,t\}$ denotes the image and text modalities respectively, and each of the VAE-GAN network consists of multiple fully connected layers. It is worth noting that $G_{*}(\cdot)$ is not only a generative network in the GAN network to generate multimodal pseudo samples from the feature space, but also a decoder in the VAE network that can reconstruct the original data according to the latent variable embedding after encoder. $G_{*}(\cdot)$ is constrained by both VAE and GAN networks during the training process, which enables the model to optimize the authenticity and diversity of generated samples at the same time, so as to improve the feature generation effect. We take image modality as an example to introduce the specific structure of our proposed multimodal feature generation.

\subsubsection{VAE Network}
In the VAE, we first input the image feature $\textbf{v}_{i}$ and its corresponding class embedding $\textbf{a}_{i}$ into the encoder $E_{v}(\cdot)$ to obtain a latent variable $z_{i}^{v} = E_{v}(\textbf{v}_{i}, \textbf{a}_{i})$. Then generator $G_{v}(\cdot)$ reconstructs the input feature $\textbf{v}_{i}$ based on the latent variable $z_{i}^{v}$ and class embedding $\textbf{a}_{i}$ to obtain the reconstruction feature $\overline{\textbf{v}}_{i} = G_{v}(z_{i}^{v}, \textbf{a}_{i})$. The objective function of this image VAE network is:
\begin{equation}
\label{equa:4-1}
\begin{aligned}
    \mathcal{L}_{VAE}^{v} &=D_{KL}(q_{\theta_{E_{v}}}(z^{v}|\textbf{v}, \textbf{a})||p(z^{v}|\textbf{a})) \\
   &-\mathbb{E}_{q_{\theta_{E_{v}}}(z^{v}|\textbf{v}, \textbf{a})}(\log p_{\theta_{G_{v}}}(\textbf{v}|z^{v}, \textbf{a})),
\end{aligned}
\end{equation}
where the former term is the Kullback-Leibler divergence loss function measuring the difference in distribution between the latent variable $z_{i}^{v}$ and the prior probability $p(z^{v}|\textbf{a})$. The $p(z^{v}|\textbf{a})$ is generally defined as the standard normal distribution. The latter term is the reconstruction loss function, which measures the element-level difference between the reconstructed image feature $\overline{\textbf{v}}$ and the original image feature $\textbf{v}$. 

\subsubsection{GAN Network}
In addition to VAE network, our CoFeG integrates a GAN network contributing to generating pseudo samples of target domain based on the class attribute information in the target domain, thereby alleviating the extreme sample size imbalance between the target and source domain. In particular, we concatenate class embedding $\textbf{a}_{i}$ with a Gaussian noise vector $z\sim \mathcal{N} (0,1)$ and input them into the generator $G_{v}(\cdot)$ for generating pseudo samples $\widetilde{\textbf{v}}_{i}=G_{v}(z_{i}, \textbf{a}_{i})$. Meanwhile, a class conditional discriminator $D_{v}(\cdot)$ is applied to determine whether the input samples ($\widetilde{\textbf{v}}_{i}$ and $\textbf{v}_{i}$) are real samples or generated samples. By the adversarial training between $D_{v}$ and $G_{v}$, the discriminator is able to provide continuous feedback to the generator, which leads to generate pseudo samples that are more consistent with the distribution of real  data. In addition, the more stable Wasserstein GAN model is used as the main architecture of the GAN network, and the objective function of the network is
\begin{equation}
\label{equa:4-3}
\begin{aligned}
	\mathcal{L}_{GAN-1}^{v}=&\mathbb{E}[D_{v}(\textbf{v}, \textbf{a}; \theta_{D_{v}}) ]-\mathbb{E}[D_{v}(\widetilde{\textbf{v}}, \textbf{a}; \theta_{D_{v}}, \theta_{G_{v}})] \\
 &-\lambda\mathbb{E}[(\|\nabla_{\hat{\textbf{v}}}D_{v}(\hat{\textbf{v}}, \textbf{a})\|_{2}-1)^{2}; \theta_{D_{v}}],
\end{aligned}
\end{equation}
where the first and second terms of the expression are  the discriminant scores of real sample $\textbf{v}$ and the generated sample $\widetilde{\textbf{v}}$, respectively. The third term is a gradient penalty, which helps to prevent pattern collapse during training by constraining the gradient change of the discriminator in the sample space. $\lambda$ is the penalty coefficient, $\nabla_{\hat{\textbf{v}}}D_{v}(\hat{\textbf{v}}, \textbf{a})$ calculates the gradient of the output of $D_{v}$ relative to the input feature $\hat{\textbf{v}}$, and the $\hat{\textbf{v}}$ is defined as
\begin{equation}
\label{equa:4-4}
\begin{matrix}
 \hat{\textbf{v}}=\epsilon\textbf{v}+(1-\epsilon)\widetilde{\textbf{v}}. & \epsilon\sim U(0,1),
\end{matrix}
\end{equation}

On the other hand, in order to make the VAE network more effective in feature reconstruction and better fit the original feature distribution pattern, the discriminator $D_{v}$ is further used to train the VAE network:
\begin{equation}
\label{equa:4-5}
\begin{aligned}
	\mathcal{L}_{GAN-2}^{v} = &\mathbb{E}[D_{v}(\textbf{v}, \textbf{a}; \theta_{D_{v}}) ]-\mathbb{E}[D_{v}(\bar{\textbf{v}}, \textbf{a}; \theta_{D_{v}}, \theta_{G_{v}}, \theta_{E_{v}})] \\
 &-\lambda\mathbb{E}[( \|\nabla_{\hat{\textbf{v}}}D_{v}(\hat{\textbf{v}}, \textbf{a}) \|_{2}-1)^{2}; \theta_{D_{v}}].
\end{aligned}
\end{equation}

By introducing a discriminator, the model can learn more detailed information about the feature distribution during the reconstruction process, thereby improving the quality of the generated samples. 

\subsubsection{Integrated Objective Function}

By combining VAE and GAN, the multimodal feature generation model proposed combines the advantages of VAE and GAN networks. On the one hand, the real feature distribution rules can be learned by encoding and reconstructing real samples. On the other hand, since the encoder's output of the latent variable $z_{i}$ and the random sampling noise $z$ conform to the same feature distribution, the generator can leverage the distribution rules learned from reconstructing the real samples, contributing to generate more realistic pseudo samples. Combined with the optimization objectives of VAE and GAN, the overall objective function of the  feature generation model is as follows:
\begin{equation}
\label{equa:4-8}
	\mathcal{L}_{VAE-GAN}^{*}=\mathcal{L}_{VAE}^{*}+\mathcal{L}_{GAN-1}^{*}+\mathcal{L}_{GAN-2}^{*},
\end{equation}

During training, the encoder $E_{*}$ and generator $G_{*}$ deceive the discriminant network by minimizing the above objective function, while the discriminator $D_{*}$ improves the discrimination ability of the generated samples by maximizing the objective function. Through adversarial training between $E_{*}$, $G_{*}$ and $D_{*}$, the discriminant network can provide more reliable feedback, which can guide the generation of network synthesis of more realistic samples. Therefore, the overall optimization goals of the image modal and text modal feature generation models are:
\begin{equation}
\label{equa:4-10}
\min_{\theta_{G_{v}}, \theta_{G_{t}}, \theta_{E_{v}}, \theta_{E_{t}}} \max_{\theta_{D_{v}}, \theta_{D_{t}}}\mathcal{L}_{VAE-GAN}^{v}+\mathcal{L}_{VAE-GAN}^{t}.
\end{equation}
where $\theta_{G_{v}}, \theta_{G_{t}}$ are the model parameters of the generator, $\theta_{E_{v}}, \theta_{E_{t}}$ are the model parameters of the encoder, and $\theta_{D_{v}}, \theta_{D_{t}}$ are the model parameters of the discriminator.

\subsection{Multimodal Feature Projection} 
\label{Common Space Projection}
Since we propose a multimodal feature generation model based on VAE-GAN composite architecture, the model can effectively generate the corresponding multimodal samples based on the target domain class $\textbf{a}\in A_{t}$ after training on the source domain entity set $X_{s}$. This section introduce the multimodal feature mapping stage.
\subsubsection{Gate Residual Network}
In the cross-modal retrieval stage, similar to previous methods, we design two projectors $P_{v}(\cdot)$ and $P_{t}(\cdot)$ to map the features of the image and text modalities into a common space. These projectors mainly consist of multiple fully connected layers, and the common space projections of image modality is as follows:
\begin{equation}
\label{equa:4-11}
\textbf{f}_{i}^{v}=P_{v}(\textbf{v}_{i}).
\end{equation}

To better utilize the knowledge extracted by vision-language pretraining model and alleviate the feature degradation, we propose a gate residual network $Gate_{*}(\cdot)$ to selectively fuse CLIP's original features and mapped features from mapper $P_{*}(\cdot)$. Specifically, we concatenate the original features $\textbf{*}_{i}$
and the mapped features $\textbf{f}_{i}^{*}$ into the gate network to obtain a gate coefficient $\textbf{g}_{i}^{*}\in\mathbb{R}^{d} $:
\begin{equation}
\label{equa:4-13}
\textbf{g}_{i}^{*}=Gate_{*}(\textbf{*}_{i}\oplus \textbf{f}_{i}^{*}),
\end{equation}
where $\oplus$ is the concatenation operation. The gate residual network outputs coefficient vectors $\textbf{g}_{i}^{*}$ according to the characteristics of the mapping features and the original features, and flexibly adjust the fusion ratio of the two features in each dimension, so as to improve the generalization ability of the few-shot multimodal projection. 

Finally, the coefficient vector $\textbf{g}_{i}^{*}$ is element-wise multiplied with the projected features and original features to obtain the final common space mapping features$\textbf{u}_{i}^{v}, \textbf{u}_{i}^{t}$:
\begin{equation}
\label{equa:4-15}
\textbf{u}_{i}^{v}=\textbf{g}_{i}^{v}\times \textbf{f}_{i}^{v} + (1-\textbf{g}_{i}^{v})\times \textbf{v}_{i},
\end{equation}
\begin{equation}
\label{equa:4-16}
\textbf{u}_{i}^{t}=\textbf{g}_{i}^{t}\times \textbf{f}_{i}^{t} + (1-\textbf{g}_{i}^{t})\times \textbf{t}_{i}.
\end{equation}

\subsubsection{Objective Function}
 \label{loss}
In order to realize the training of the above multimodal feature projecting model, three different objective loss functions are proposed: classification loss, modal consistency loss, and contrastive learning loss. First, since class labels are important for common space projection, we use a linear mapping layer with a Softmax activation function, based on the common space mapping feature $\textbf{u}_{i}^{*}$ and class label of predicted samples $\hat{\textbf{y}}^{*}_{i}$, and compute the cross-entropy loss function:
\begin{equation}\label{equa:18}
\mathcal{L}_{1}=-\frac{1}{n} \sum_{i=1}^{n}\sum_{c=1}^{C}\left ( {\textbf{y}_{ic}}\log\left ( \hat{\textbf{y}}_{ic}^{v} \right )+{\textbf{y}_{ic}}\log\left ( \hat{\textbf{y}}_{ic}^{t} \right ) \right ),
\end{equation}
where $C$ is the number of class, $n$ is the overall number of the real samples and the generated samples during training.

Second, the pairwise relations between modalities plays an important role in bridging the heterogeneity gap, and the distance between paired cross-modal samples in the common space should be as small as possible. Hence, we design a modal consistency loss:
\begin{equation}\label{equa:18}
\mathcal{L}_{2}=-\frac{1}{n}\sum_{i=1}^{n}||\textbf{u}_{i}^{v}-\textbf{u}_{i}^{t}||_{2}.
\end{equation}

Finally, we introduce an instance-level contrast learning loss function to pull in pairwise cross-modal samples and push away unpaired samples. The cross-modal comparison learning loss is:
\begin{equation}\label{equa:19}
\mathcal{L}_{3}=-\frac{1}{n} \sum_{*_{i}}^{v,t}\sum_{i=1}^{n}\log(P(\textbf{u}_{i}^{*_{i}})),
\end{equation}
where $P(\textbf{u}_{i}^{*_{i}})$ is the matching similarity between $\textbf{u}_{i}^{*_{i}}$ and its paired entities. Specifically, we calculate the similarity between $\textbf{u}_{i}^{*_{i}}$ and all other entities in the same training batch and normalize:
\begin{equation}\label{equa:19}
P(\textbf{u}_{i}^{*_{i}})=\frac{\exp(\cos(\textbf{u}_{i}^{v}, \textbf{u}_{i}^{t})/\tau)}{\sum_{*_{i},*_{j}}^{v,t}\sum^{n}_{j=1}\exp(\cos(\textbf{u}_{j}^{*_{j}}, \textbf{u}_{i}^{*_{i}})/\tau)},
\end{equation}
where $*_{i}, *_{j}$ are the modal categories of samples $\textbf{u}_{i}^{*_{i}}, \textbf{u}_{j}^{*_{j}}$, $\cos(\cdot, \cdot)$ is the cosine similarity function between the two eigenvectors, and $\tau$ is a temperature coefficient that adjusts the model's attention to pairs of cross-modal samples.

In summary, in the multimodal feature mapping stage, the overall optimization of the model are:
\begin{equation}\label{equa:21}
\mathcal{L}=\alpha\mathcal{L}_{1}+\beta\mathcal{L}_{2}+\gamma\mathcal{L}_{3},
\end{equation}
where $\alpha, \beta, \gamma$ are a set of hyper-parameters that are used to balance the weights of different optimization goals during model training.

\section{Experiments}
\subsection{Datasets}
We evaluate our model on image-text retrieval, which includes four widely used benchmark datasets: Wikipedia \cite{rasiwasia2010}, Pascal Sentence \cite{rashtchian2010collecting}, NUS-WIDE \cite{chua2009nus}, and NUS-WIDE-10K \cite{chua2009nus}.
\begin{itemize}
    \item \textbf{Wikipedia} has 2,866 instances of image–text pairs that are crawled from the Wikipedia website, and each instance has one label from ten categories. In order to ensure the rationality of the experiment, 2,173 and 693 pairs of image-text entities are randomly selected from the dataset as the original training set and the original test set, respectively.
    \item \textbf{Pascal Sentence }consists of 1,000 images, with each image belonging to one of the predefined 20 categories. Each image is also described in five textual sentences. In our experiment, 800 and 200 pairs of image-text entities are randomly selected from the dataset as the original training set and the original test set, respectively.
    \item \textbf{NUS-WIDE} contains 270,000 images with associated tags, where images belong to at least one of the 81 categories. We refer to the previous method \cite{chi2018dual} to extract 42,859 pairs of image-text entities from the 10 categories with the highest occurrence frequency of the dataset to form the NUS-WIDE dataset. Its original training set and original test set contained 25,685 and 12,174 pairs of samples, respectively.
    \item \textbf{Nuswide-10k} is a simplified version of NUS-WIDE, which contains 10,000 pairs image-text instance. In this work, 1,000 pairs of image-text entities are randomly selected from the 10 most frequent categories of the NUS-WIDE dataset to form the NUS-WIDE-10K dataset by referring to the previous method \cite{nuswide10k}. Its original training set and original test set contained 8,000 and 2,000 pairs of samples, respectively.
\end{itemize}

\begin{figure}
    \centering
    \includegraphics[width=0.75\linewidth]{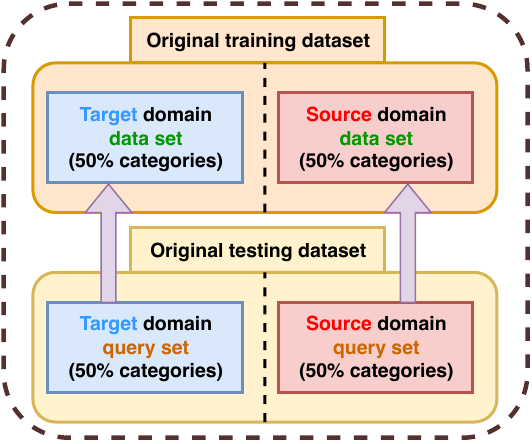}
    \caption{The illustration of the dataset splitting setting.}
    \label{fig:table1}
\end{figure}

In the few-shot scenario, the original dataset needs to be divided in more detail because the data categories of the source domain and the target domain are different. We further divide the dataset into four subsets: Target Data Set, Source Data Set, Target Query Set, and Source Query Set. The target domain and the source domain each contain 50\% of the data categories, as shown in Figure \ref{fig:table1}. In the training stage, the model is trained on the source domain set. In the validation phase, the model takes the data in the source domain query set as a query and retrieves the data in the source domain sample set, and in the testing stage, the model extracts the data from the target domain query set as a query and retrieves the data in the target domain sample set. The statistics of the four datasets are shown in the table \ref{tab:1}.

\begin{table}
    \centering
    \caption{The statistics of the four datasets. ``xx/xx'' denotes ``seen/unseen'' in ``class'' or ``image/text'' in ``Train'' and ``Test''.}
    \label{tab:1}
    \scalebox{0.9}{
    \begin{tabular}{c|cccc}
        \hline \hline
         Datasets & Pairs & Classes  & Train & Test \\ \hline
         Wikipedia & 2,866 & 5/5 & 2,173/2,173 & 693/693 \\
         Pascal Sentence & 1,000 & 10/10 & 800/800 & 200/200 \\
         NUSWIDE & 42,859 & 5/5 & 25,685/25,685 & 12,174/12,174 \\
         Nuswide-10k & 10,000 & 5/5 & 8,000/8,000 & 2,000/2,000 \\ \hline \hline
    \end{tabular}}
    
\end{table}

\subsection{Evaluation Metric and Baseline Models}
For the evaluation metric, we calculate the mean average precision (MAP) score for image-to-text retrieval (using image queries) and text-to-image retrieval (using text queries). We first calculate the average precision (AP) :
\begin{equation}
    AP = \frac{1}{T} \sum_{r=1}^R P_r \times \delta(r),
\end{equation}
where $R$ is the set of retrieved items, $T$ is the number of relevant items in the retrieved items, $P(r)$ denotes the precision of the top $r$ retrieved items, and $\delta (r)$ is an indicator function. We adopt the cosine distance to measure the similarity of features. If the $r$-th item is relevant to the query, the value of $\delta (r)$ is $1$. The MAP is the mean value of the AP score, which jointly considers the ranking information and precision:
\begin{equation}\label{equa:21}
\textrm{mAP}=\frac{1}{n}\sum_{i=1}^{n}\textrm{AP}(i),
\end{equation}
where $n$ is the number of the query.

To evaluate our model, we conduct experiments on image-to-text (Img2Txt) and text-to-image (Txt2Img) tasks under the zero-shot and few-shot scenarios. Particularly, the latter scenario includes 1-shot, 3-shot, and 5-shot. We compare our proposed FLEX-CLIP method with 10 state-of-the-art methods on the retrieval tasks, including CCA \cite{Hotelling1992}, DSCMR \cite{zhen2019deep}, MARS \cite{wang2022Mars}, MRL \cite{Hu2021Mrl}, DADN \cite{chi2019zero}, CFSA \cite{xu2020correlated}, JSFE \cite{xu2022joint}, MDVAE \cite{tian2022multimodal}, CLIP \cite{radford2021learning}. We reimplement these models and the experimental results are shown in Table \ref{tab:3}. We briefly describe the baseline models as follows:
\begin{itemize}
    \item \textbf{DSCMR \cite{zhen2019deep}} aims to find a common representation space, minimizes the discrimination loss in both the label space and the common representation space, and minimizes the modality loss with a weight-sharing strategy.
    \item \textbf{MARS \cite{wang2022Mars}} treats the label information as a distinct modality and introduces a label parsing module, LabNet, to generate a semantic representation for correlating different modalities.
    \item \textbf{MRL \cite{Hu2021Mrl}} tries to learn with multimodal noisy labels to mitigate noisy samples and correlate distinct modalities simultaneously, consisting of a Robust Clustering loss (RC) and a Multimodal Contrastive loss (MC).
    \item \textbf{DADN \cite{chi2019zero}} learns common embeddings and explores the knowledge from word-embeddings of categories, in which zero-shot cross-media dual generative adversarial networks architecture is proposed, consisting of two kinds of generative GANs for common embedding generation.
    \item \textbf{CFSA \cite{xu2020correlated}} integrates multimodal feature synthesis by utilizing class-level word embeddings to guide two coupled Wasserstein generative adversarial networks (WGANs) to synthesize sufficient multimodal features.
    \item \textbf{MDVAE \cite{tian2022multimodal}} is a novel zero-shot cross-modal retrieval model, which consists of two coupled disentanglement variational autoencoders (DVAEs) and a fusion-exchange VAE (FVAE).
    \item \textbf{JFSE \cite{xu2022joint}} jointly performs multimodal feature synthesis and common embedding space learning, consisting of two coupled conditional Wasserstein GAN modules for two modalities and three advanced distribution alignment schemes.
    \item \textbf{CLIP\cite{radford2021learning}} is a neural network trained on a variety of (image, text) pairs. It can be instructed in natural language to predict the most relevant text snippet, given an image, without directly optimizing for the task.
\end{itemize}

\begin{table}[t]
    \centering
    \caption{Relevant experimental parameters of different datasets. The LR denotes the learning rate.}
    \label{tab:2}
    \scalebox{0.9}{
    \begin{tabular}{c|ccc|cc}
    \hline \hline
         \multirow{2}{*}{Dataset}&\multicolumn{3}{c|}{Multimodal Feature Generation}& \multicolumn{2}{c}{Common Space Projection}\\ \cline{2-6}
         &  Batch&LR&Gen Num& Batch& LR\\ \hline
         Wikipedia &  256&$1\times 10^{-3}$ &70& 256 & $1\times10^{-3}$\\
         Pascal Sentence &  64&$1\times 10^{-3}$ &30& 64 & $4\times 10^{-4}$\\
         NUS-WIDE &  512&$2\times 10^{-3}$ &500& 512 & $4\times 10^{-4}$\\
         NUS-WIDE-10K &  2,048&$1\times 10^{-3}$ &300& 2,048& $5\times 10^{-3}$\\ \hline \hline
    \end{tabular}}
\end{table}

\subsection{Implementation Details}
In our network architecture, the encoders $E_{*}$, comprise three fully-connected layers, progressively configured with dimensions [1,024,800,512]. Each of these layers employs the ReLU activation function, while the last layer utilizes the Sigmoid activation for smoother output representation. The generators $G_{*}$, are constructed using three fully-connected layers with respective dimensional configurations of [1,536,800,1,024]. Like the encoders, every layer within these generators utilizes the ReLU activation function, and the final layer adopts the Sigmoid activation function to ensure optimized outputs. As for the adversarial learning component, we architect the discriminators $D_{*}$, with two fully-connected layers [2,048,1] units respectively. The intermediate layer of this discriminator harnesses the LeakyReLU activation function to enhance its discriminative capacity.

\begin{table*}[t]
\centering
    \label{tab:3}
    \caption{Comparisons of our approach and 9 compared methods on four benchmark datasets for zero-shot image-text retrieval. The best and second-best results are marked in bold and underlined font, respectively.}
    \resizebox{\linewidth}{!}{
\begin{tabular}{c|ccc|ccc|ccc|ccc}
\hline \hline
\multirow{2}{*}{Model} & \multicolumn{3}{l}{Wikipedia} & \multicolumn{3}{l}{Nuswide-10k} & \multicolumn{3}{l}{Pascal Sentence} & \multicolumn{3}{l}{NUSWIDE} \\
                       & Img2Txt   & Txt2Img  & Avg    & Img2Txt   & Txt2Img   & Avg     & Img2Txt     & Txt2Img    & Avg      & Img2Txt  & Txt2Img  & Avg   \\ \hline
CCA                    & 0.359     & 0.306    & 0.332  & 0.320     & 0.323     & 0.322   & 0.272       & 0.233      & 0.252    & 0.411    & 0.408    & 0.410 \\ \hline
DSCMR         & 0.362     & 0.336    & 0.349  & 0.346     & 0.377     & 0.361   & 0.455       & 0.448      & 0.452    & 0.411    & 0.431    & 0.421 \\ \hline
MARS          & 0.349     & 0.318    & 0.333  & 0.280     & 0.292     & 0.286   & 0.385       & 0.384      & 0.384    & 0.339    & 0.367    & 0.353 \\
MRL            & 0.418     & 0.455    & 0.437  & 0.454     & 0.452     & 0.453   & 0.519       & 0.503      & 0.511    & 0.498    & 0.491    & 0.494 \\ \hline
DADN          & 0.369     & 0.325    & 0.347  & 0.342     & 0.355     & 0.349   & 0.442       & 0.398      & 0.420    & 0.417    & 0.443    & 0.430 \\
CFSA          & 0.421     & 0.356    & 0.388  & 0.383     & 0.402     & 0.392   & 0.463       & 0.440      & 0.451    & 0.450    & 0.473    & 0.461 \\
JSFE          & 0.420     & 0.352    & 0.386  & 0.379     & 0.403     & 0.391   & 0.438       & 0.423      & 0.431    & 0.458    & 0.473    & 0.465 \\
MDVAE         & 0.474     & 0.459    & 0.466  & 0.400     & 0.425     & 0.412   & 0.470       & 0.467      & 0.469    & 0.399    & 0.424    & 0.411 \\ \hline
CLIP           & \underline{0.526}     & \underline{0.479}    & \underline{0.502}  & \underline{0.480}     & \underline{0.526}     & \underline{0.503}   & \underline{0.669}       & \underline{0.659}      & \underline{0.664}    & \underline{0.581}    & \underline{0.602}    & \underline{0.592} \\ \hline
OURS                   & \textbf{0.539}     & \textbf{0.512}    & \textbf{0.526}  & \textbf{0.569}     & \textbf{0.589}     & \textbf{0.579}   & \textbf{0.690}       & \textbf{0.689}      & \textbf{0.689 }   & \textbf{0.661}    & \textbf{0.680}    & \textbf{0.671} \\ \hline \hline
\end{tabular}}
\end{table*}

We conduct two training stages. In the first phase, the multimodal feature generation model is first trained with source domain data and a small amount of target domain data. In the second stage, the trained generative model is used to generate a specific number of target domain pseudo-entities. Subsequently, the generated pseudo-entities are combined with the source domain dataset to train the multimodal mapping model. In the training process, the Adam optimizer is used for training optimization. Table \ref{tab:2} shows the training parameters of the FLEX-CLIP model on different datasets. 

\subsection{Comparison Results}
\subsubsection{Zero-shot Scenario}
Zero-shot learning represents an extreme form of few-shot learning, where the training data contains no samples from the target domain. We conduct zero-shot cross-modal retrieval experiments on the FLEX-CLIP model and baseline models, with the results presented in Table 1. The key observations are as follows:

\paragraph{FLEX-CLIP outperforms other methods significantly}. The zero-shot setting challenges these methods in handling unseen classes in the target domain. However, our approach alleviates the issues of \textbf{\textit{extreme data imbalance}} and \textbf{\textit{domain knowledge bias}} in zero-shot scenarios by synthesizing pseudo-samples in the target domain via our composite VAE-GAN generation network and fully leveraging the pre-trained semantic features of images and texts.



\paragraph{Zero-shot methods perform better.}Compared with conventional methods (CCA, DSCMR, MARS), zero-shot cross-modal methods (DADN, CFSA, JSFE, and MDVAE) have certain advantages in most cases. But compared with the zero-shot methods, the robust method MRL shows competitive performance on all four datasets. Indicating that with the help of the high-quality original features extracted by the vision-language pretraining model, the robust learning architecture of MRL \textbf{can better capture the deep semantic association between cross-modal data, so that it can have better performance in zero-shot cross-modal retrieval tasks. }

\paragraph{FLEX-CLIP solve the feature degradation.} However, all the baseline models suffer from \textbf{\textit{feature degradation}} in the zero-shot scenario, and the zero-shot cross-modal retrieval performance of the model is significantly weaker than that of the CLIP model. FLEX-CLIP is the model that surpasses the original CLIP features. Specificly, compared with the best counterpart CLIP, our approach achieves notable improvement and consistently beats CLIP with 2.31, 7.59, 2.53, and 7.90 improvements of the average MAP scores on four datasets. It demonstrates that our FLEX-CLIP better utilizes the multimodal knowledge extracted by vision-language pre-trained model and greatly solve the \textbf{\textit{feature degradation}} by the proposed gate residual network. We will further explore the improvement on CLIP in our CASE STUDY.

\subsubsection{Few-shot Scenario}
We conduct few-shot cross-modal retrieval experiments by randomly selecting 1,3,5-shot, shown in Table 2, leading to the following observations:

\begin{table*}[]
\caption{Comparisons of our approach and 9 compared methods on four benchmark datasets for 1,3,5-shot image-text retrieval. The 1,3,5-shot experiment results are same in CLIP, so the sample information is ``-''.}
    \label{tab:4}
    \resizebox{\linewidth}{!}{
\begin{tabular}{c|c|ccc|ccc|ccc|ccc}
\hline \hline
\multicolumn{2}{l}{\multirow{2}{*}{Model}} & \multicolumn{3}{l}{Wikipedia} & \multicolumn{3}{l}{Nuswide-10k} & \multicolumn{3}{l}{Pascal Sentence} & \multicolumn{3}{l}{NUSWIDE} \\
\multicolumn{2}{l}{}                       & I2T   & T2I  & Avg    & I2T   & T2I   & Avg     & I2T     & T2I    & Avg      & I2T  & T2I  & Avg   \\ \hline
\multirow{3}{*}{CCA}             & 1-shot  & 0.357     & 0.306    & 0.331  & 0.321     & 0.323     & 0.322   & 0.251       & 0.239      & 0.245    & 0.411    & 0.409    & 0.410 \\ 
                                 & 3-shot  & 0.360     & 0.309    & 0.334  & 0.322     & 0.324     & 0.323   & 0.312       & 0.305      & 0.309    & 0.412    & 0.409    & 0.410 \\
                                 & 5-shot  & 0.358     & 0.311    & 0.334  & 0.323     & 0.325     & 0.324   & 0.337       & 0.351      & 0.344    & 0.412    & 0.407    & 0.409 \\ \hline
\multirow{3}{*}{DSCMR}   & 1-shot  & 0.360     & 0.331    & 0.346  & 0.340     & 0.375     & 0.358   & 0.443       & 0.459      & 0.451    & 0.416    &  0.427        & 0.422      \\
                                 & 3-shot  & 0.361     & 0.338    & 0.349  & 0.336     & 0.369     & 0.353   & 0.446       & 0.477      & 0.462    & 0.420    & 0.426    & 0.423 \\
                                 & 5-shot  & 0.354     & 0.332    & 0.343  & 0.339     & 0.373     & 0.356   & 0.534       & 0.569      & 0.552    & 0.412    & 0.434    & 0.423 \\ \hline
\multirow{3}{*}{MARS}   & 1-shot  & 0.330     & 0.308    & 0.319  & 0.279     & 0.289     & 0.284   & 0.391       & 0.385      & 0.388    & 0.348    & 0.374    & 0.361 \\
                                 & 3-shot  & 0.328     & 0.311    & 0.320  & 0.279     & 0.288     & 0.283   & 0.452       & 0.460      & 0.456    & 0.341    & 0.386    & 0.364 \\
                                 & 5-shot  & 0.342     & 0.361    & 0.351  & 0.283     & 0.294     & 0.288   & 0.551       & 0.585      & 0.568    & 0.351    & 0.376    & 0.363 \\ \hline
\multirow{3}{*}{MRL}     & 1-shot  & 0.427     & 0.461    & 0.444  & 0.459     & 0.449     & 0.454   & 0.529       & 0.517      & 0.523    & 0.492    & 0.491    & 0.491 \\
                                 & 3-shot  & 0.433     & 0.475    & 0.454  & 0.470     & 0.457     & 0.464   & 0.566       & 0.549      & 0.557    & 0.506    & 0.503    & 0.504 \\
                                 & 5-shot  & 0.412     & 0.480    & 0.446  & 0.471     & 0.457     & 0.464   & 0.619       & 0.610      & 0.615    & 0.509    & 0.489    & 0.499 \\ \hline
\multirow{3}{*}{DADN}   & 1-shot  & 0.358     & 0.320    & 0.339  & 0.340     & 0.354     & 0.347   & 0.457       & 0.382      & 0.420    & 0.412    & 0.441    & 0.426 \\
                                 & 3-shot  & 0.358     & 0.333    & 0.346  & 0.344     & 0.355     & 0.350   & 0.457       & 0.424      & 0.440    & 0.414    & 0.439    & 0.426 \\
                                 & 5-shot  & 0.352     & 0.328    & 0.340  & 0.340     & 0.353     & 0.347   & 0.482       & 0.553      & 0.518    & 0.409    & 0.437    & 0.423 \\ \hline
\multirow{3}{*}{CFSA}   & 1-shot  & 0.400     & 0.348    & 0.374  & 0.381     & 0.403     & 0.392   & 0.495       & 0.491      & 0.493    & 0.451    & 0.471    & 0.461 \\
                                 & 3-shot  & 0.409     & 0.349    & 0.379  & 0.371     & 0.393     & 0.382   & 0.577       & 0.578      & 0.578    & 0.450    & 0.470    & 0.460 \\
                                 & 5-shot  & 0.404     & 0.352    & 0.378  & 0.378     & 0.398     & 0.388   & 0.636       & 0.654      & 0.645    & 0.458    & 0.477    & 0.467 \\ \hline
\multirow{3}{*}{JSFE}   & 1-shot  & 0.405     & 0.349    & 0.377  & 0.380     & 0.407     & 0.394   & 0.523       & 0.517      & 0.520    & 0.447    & 0.467    & 0.457 \\
                                 & 3-shot  & 0.393     & 0.337    & 0.365  & 0.382     & 0.399     & 0.390   & 0.578       & 0.577      & 0.577    & 0.450    & 0.469    & 0.460 \\
                                 & 5-shot  & 0.414     & 0.350    & 0.382  & 0.378     & 0.399     & 0.388   & 0.622       & 0.647      & 0.635    & 0.447    & 0.464    & 0.456 \\ \hline
\multirow{3}{*}{MDVAE}  & 1-shot  & 0.450     & 0.433    & 0.442  & 0.418     & 0.437     & 0.428   & 0.448       & 0.455      & 0.451    & 0.397    & 0.419    & 0.408 \\
                                 & 3-shot  & 0.476     & 0.445    & 0.460  & 0.403     & 0.434     & 0.418   & 0.491       & 0.489      & 0.490    & 0.396    & 0.410    & 0.403 \\ 
                                 & 5-shot  & 0.497     & 0.450    & 0.473  & 0.410     & 0.444     & 0.427   & 0.525       & 0.508      & 0.516    & 0.407    & 0.418    & 0.413 \\ \hline
\multirow{1}{*}{CLIP}  & - & \underline{0.526}     & \underline{0.479}    & \underline{0.502}   & \underline{0.480}      & \underline{0.526}      & \underline{0.503}    & \underline{0.669}        & \underline{0.659}       & \underline{0.664}     & \underline{0.581}     & \underline{0.602}    & \underline{0.592}  \\ \hline
\multirow{3}{*}{Ours(1-shot)}    & 1-shot  & \textbf{0.551}     & \textbf{0.523}    & \textbf{0.537}  & \textbf{0.573 }    & \textbf{0.592}     & \textbf{0.582}   & \textbf{0.695}       & \textbf{0.690}      & \textbf{0.693}    & \textbf{0.660}    & \textbf{0.677}    & \textbf{0.668} \\
                                 & 3-shot  & \textbf{0.594}     & \textbf{0.554}    & \textbf{0.574}  & \textbf{0.609}     & \textbf{0.626}     & \textbf{0.618}   & \textbf{0.713}       & \textbf{0.718}      & \textbf{0.715}    & \textbf{0.666}    & \textbf{0.687}    & \textbf{0.676} \\
                                 & 5-shot  & \textbf{0.565}     & \textbf{0.533}    & \textbf{0.549}  & \textbf{0.585}     & \textbf{0.603}     & \textbf{0.594}   & \textbf{0.701}       & \textbf{0.700}      & \textbf{0.700}    & \textbf{0.662}    & \textbf{0.680}    & \textbf{0.671} \\ \hline \hline
\end{tabular}}
\end{table*}



\paragraph{Noise data and outliers problem:} From the experimental results, it can be seen that almost all methods show a certain performance improvement on the Pascal Sentence and Wikipedia datasets with the increase of the number of target domain samples. This indicates that the increase in the sample size can help alleviate the extreme imbalance between the sample size of the target domain and the source domain in the few-sample scenario. However, on the NUS-WIDE-10K and NUS-WIDE datasets, the performance of most of the methods becomes unstable, instead of improving significantly with the increase of the number of samples in the target domain. This may be due to the larger amount of data and more noise data for the NUS-WIDE-10K and NUS-WIDE. A small number of randomly selected samples of the target domain introduced by few-shot setting may contain outliers that cause the model to deviate from the actual target when it is trained.

\paragraph{FLEX-CLIP's adaptability to noise in few-shot.} It is worth noting that with the increase of the number of samples in the target domain, the FLEX-CLIP model achieves performance improvement on all four datasets. This indicates that the proposed composite VAE-GAN architecture of FLEX-CLIP can better capture the feature distribution pattern of the target domain from a small number of samples, so as to \textbf{alleviate the influence of noise nodes} on model training. Based on this learning ability to capture the feature distribution pattern, the feature generation model can alleviate the \textbf{\textit{data imbalance problem}} faced by the downstream mapping model by generating the target domain pseudo-entity.

\begin{figure*}[t]
    \centering
    \includegraphics[width=1\linewidth]{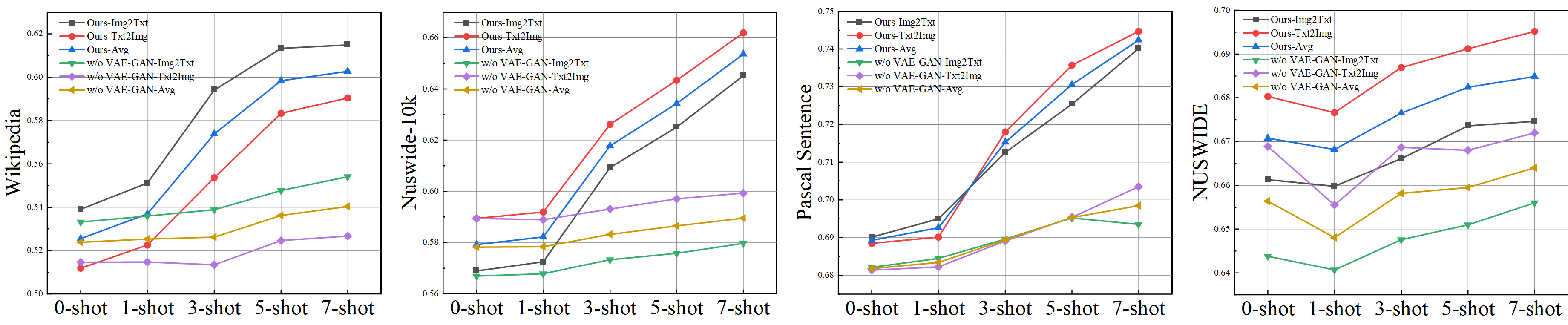}
    \caption{The mAP curve of the ablation experiment of the composite VAE-GAN network.}
    \label{fig:table5}
\end{figure*}

\begin{table*}[t]
    \label{tab:7}
    \caption{The effectiveness of the ablation experiment on the residual gate network. The $L_1$ is the cross-entropy loss function, $L_2$ is a modal consistency loss function, and $L_3$ is cross-modal comparison learning loss, as detailedly described in Sec. \ref{loss}. }
    \resizebox{\linewidth}{!}{
\begin{tabular}{c|ccc|ccc|ccc|ccc}
\hline \hline
\multirow{2}{*}{Model} & \multicolumn{3}{c}{Wikipedia} & \multicolumn{3}{c}{Nuswide-10k} & \multicolumn{3}{c}{Pascal Sentence} & \multicolumn{3}{c}{NUSWIDE} \\
                       & Img2Txt  & Txt2Img  & Avg     & Img2Txt   & Txt2Img   & Avg     & Img2Txt    & Txt2Img    & Avg       & Img2Txt  & Txt2Img & Avg    \\ \hline
Ours(0-shot)          &\textbf{ 0.5391 }  & \textbf{0.5119}   & \textbf{0.5255}  & \textbf{0.5689 }   & \textbf{0.5894}    & \textbf{0.5792}  & \textbf{0.6901}     & \textbf{0.6885}     & \textbf{0.6893}    & \textbf{0.6613}   & \textbf{0.6803}  & \textbf{0.6708} \\
w/o Res-Gate           & 0.4790   & 0.4462   & 0.4626  & 0.5426    & 0.5651    & 0.5539  & 0.5366     & 0.5365     & 0.5366    & 0.6510   & 0.6640  & 0.6575 \\
w/o L1                 & 0.5303   & 0.5043   & 0.5173  & 0.5669    & 0.5846    & 0.5758  & 0.6880     & 0.6858     & 0.6869    & 0.6428   & 0.6592  & 0.6510 \\
w/o L2                 & 0.5389   & 0.5109   & 0.5249  & 0.5664    & 0.5866    & 0.5765  & 0.6819     & 0.6807     & 0.6813    & 0.6356   & 0.6599  & 0.6478 \\
w/o L3                 & 0.4865   & 0.4542   & 0.4704  & 0.4691    & 0.4798    & 0.4745  & 0.6812     & 0.6739     & 0.6776    & 0.5352   & 0.5241  & 0.5297 \\\hline \hline
\end{tabular}}
\end{table*}

\subsubsection{Case Study}
\begin{figure}
    \centering
    \includegraphics[width=1.0\linewidth]{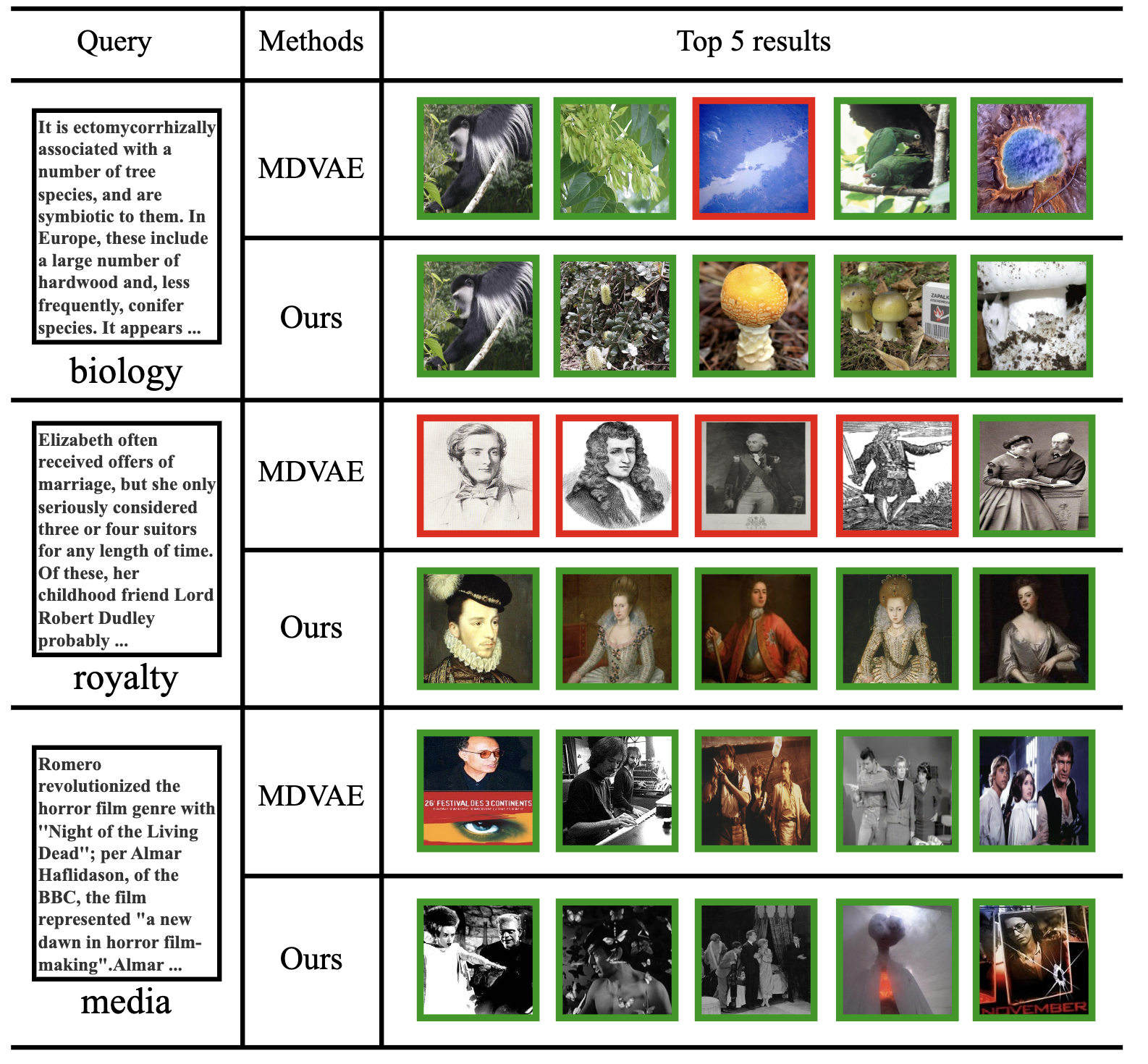}
    \caption{The examples of top-5 retrieval results of our FLEX-CLIP and the compared model MDVAE.}
    \label{fig:4}
\end{figure}

In addition, we also provide typical zero-shot image-text retrieval exemplars obtained by our FLEX-CLIP and compared model MDVAE, shown in Figure \ref{fig:4}. For each query text, its top 5 retrieved images are presented, the red rectangles denote the wrong retrieved candidates, and the green rectangles denote the correctly retrieved candidates. It can be seen that our method successfully retrieves the correct images with text queries in most cases, while the MDVAE sometimes retrieves the wrong candidates especially for some image samples that are very similar in appearance. For example, in the second example of ``royalty'' retrieval, since the similar appearances of different Medieval portraits, MDVAE mistakes other portraits for Elizabeth and her associates, and our model FLEX-CLIP correctly identified the relevant candidates.

The results demonstrate that our model FLEX-CLIP is advanced to retrieve the semantically related images with high accuracy, which indicates that our FLEX-CLIP learns effective multimodal semantic features for correlated images and texts, alleviate the data imbalance and feature degradation, and improve the zero-shot cross-modal retrieval ability.


\subsection{Ablation Study}
\subsubsection{Effect of Generative VAE-GAN Network}
\textbf{\textit{To validate the effectiveness of our network for generating multimodal features to alleviate the data imbalance}}, we perform ablation experiments under 0-shot, 1-shot, 3-shot, 5-shot, and 7-shot scenarios which compare the w/o VAE-GAN generation network FLEX-CLIP model with the standard FLEX-CLIP. We conduct experiments on four datasets, and the complete experimental results of Img2Txt and Txt2Img tasks are shown in Figure 3. 

We can observe that compared with our standard model, the model performance of the w/o VAE-GAN composite network has a significant decrease in most scenarios, which proves that our proposed composite generation network, utilizing the feature generation method, can effectively solve the problem of \textbf{extreme data imbalance} in the zero-shot setting. Notably, the results of the w/o VAE-GAN model are better than CLIP, indicating that the gate network improves the performance of the origin CLIP features. 

In addition, in the absence of the VAE-GAN composite network, it can be observed that the model's effect improves significantly more slowly with the increase in the number of samples compared to the standard model. This proves that our model also has a good ability to learn under few-shot and can fully utilize the information of very few target domain samples to solve the data imbalance problem.

\begin{table*}[t]
\caption{The effectiveness of the ablation experiment on the VAE network.}
\label{tab:6}
\resizebox{\linewidth}{!}{
\begin{tabular}{c|ccc|ccc|ccc|ccc}
\hline  \hline 
\multirow{2}{*}{Model} & \multicolumn{3}{|c|}{Wikipedia} & \multicolumn{3}{c|}{Nuswide-10k} & \multicolumn{3}{c|}{Pascal Sentence} & \multicolumn{3}{c}{NUSWIDE} \\
                       & Img2Txt  & Txt2Img  & Avg     & Img2Txt   & Txt2Img   & Avg     & Img2Txt    & Txt2Img    & Avg       & Img2Txt  & Txt2Img & Avg    \\ \hline 
Ours(0-shot)           & 0.5391   & 0.5119   & 0.5255  & 0.5689    & 0.5894    & 0.5792  & 0.6901     & 0.6885     & 0.6893    & 0.6613   & 0.6803  & 0.6708 \\
Ours(1-shot)           & 0.5512   & 0.5226   & 0.5369  & 0.5725    & 0.5919    & 0.5822  & 0.6950     & 0.6901     & 0.6926    & 0.6598   & 0.6766  & 0.6682 \\
Ours(3-shot)           & 0.5941   & 0.5536   & 0.5738  & 0.6093    & 0.6262    & 0.6178  & 0.7126     & 0.7180     & 0.7153    & 0.6661   & 0.6869  & 0.6765 \\
Ours(5-shot)           & 0.6133   & 0.5834   & 0.5984  & 0.6251    & 0.6434    & 0.6343  & 0.7254     & 0.7357     & 0.7306    & 0.6736   & 0.6912  & 0.6824 \\
w/o VAE(0-shot)        & 0.5382   & 0.5122   & 0.5252  & 0.5666    & 0.5869    & 0.5768  & 0.6810     & 0.6829     & 0.6820    & 0.6560   & 0.6731  & 0.6646 \\
w/o VAE(1-shot)        & 0.5404   & 0.5155   & 0.5280  & 0.5681    & 0.5880    & 0.5781  & 0.6890     & 0.6892     & 0.6891    & 0.6484   & 0.6696  & 0.6590 \\
w/o VAE(3-shot)        & 0.5440   & 0.5211   & 0.5326  & 0.5688    & 0.5904    & 0.5796  & 0.6899     & 0.6935     & 0.6917    & 0.6585   & 0.6771  & 0.6678 \\
w/o VAE(5-shot)        & 0.5533   & 0.5266   & 0.5400  & 0.5703    & 0.5911    & 0.5807  & 0.6950     & 0.6978     & 0.6964    & 0.6618   & 0.6795  & 0.6707 \\ \hline \hline 
\end{tabular}}
\end{table*}

\subsubsection{Effect of VAE Network on Few-shot}
\textbf{\textit{To verify the effect of the VAE network on capturing the feature distribution to better generation}}, we perform the ablation experiments on the four datasets. 
We perform ablation experiments under 0-shot, 1-shot, 3-shot, and 5-shot scenarios, and the experimental results of the Img2Txt, Txt2Img, and the average scores are shown in Table \ref{tab:6}.

We can observe that compared to the results of the model without VAE, our standard basic model shows a significant improvement in different sample settings, especially in the 5-shot scenario, demonstrating an improvement of 5.84, 5.36, 3.42, and 1.17 on the four datasets respectively. In addition, after removing the VAE network, the model performance improvement due to the increase of training samples in the target domain is relatively limited. This proves that the introduction of VAE greatly helps the model to capture the correct feature distribution patterns, which can effectively improve the feature generation network to solve the problem of \textbf{data imbalance} in the few-shot scenario. The above results demonstrate that the introduction of VAE effectively improves the performance of feature generation and further enhances the effectiveness of X-shot cross-modal retrieval.

\subsubsection{Effect of Residual Gate Network and Loss Function on Common Space Mapping}
\textbf{\textit{To verify the effect of the residual gate network network and every loss function on alleviating the feature degradation on the target domain,}} we compare the effect of the proposed residual gate network and three loss functions as well as our proposed standard scheme in our FLEX-CLIP. 

We take the zero-shot scenario as a testbed and report the average retrieval performance and the detailed results of the four datasets without residual gate network, and three loss functions as in Table 4. We can observe that using the proposed residual gate network in FLEX-CLIP obtains better performance on all the datasets than the scheme without a gate network, demonstrating that the selective fusion of the image and text features extracted by CLIP can greatly alleviate the \textbf{feature degradation} problem. Moreover, we can observe that the retrieval results of the three schemas, respectively without $L_1$, $L_2$, and $L_3$, the cross-entropy loss function, modal consistency loss function, cross-modal comparison learning loss, respectively, all show a decrease in the MAP scores on both Img2Txt and Txt2Img task. Specifically, without the cross-modal comparison learning loss ($L_3$), the model presents a more significant decrease than the other two ablation models, indicating that the measuring distance between each sample and the other samples is more effective than directly modeling the distance between paired cross-modal samples, which contributes more to the bridge the heterogeneity multimodal gap in solving feature degradation.


\section{Conclusion}
In this paper, we propose \textbf{F}eature-\textbf{L}evel G\textbf{E}neration Network to enhance CLIP features for \textbf{X}-shot CMR, a novel approach that performs feature generation. In multimodal feature generation, we design a composite multimodal VAE-GAN network, which combines the advantages of VAE and GAN to solve the data imbalance issue in few-shot cross-modal retrieval. Notably, the introduced VAE structure effectively improves the feature distribution of the generated features. In common space projection, 
by selectively fusing the original features extracted by CLIP and the features projected into a common space, our model fully utilizes the multimodal semantic information from the vision-language pretraining model. We have conducted several comprehensive experiments on four benchmark datasets, demonstrating the superiority of our proposed FLEX-CLIP method on both image-to-text and text-to-image cross-modal retrieval tasks and our results outperform the state-of-the-art. 
In future work, we attempt to improve the feature generation methods to generate features that can be generalized to more task scenarios in the multimodal community.

\bibliographystyle{ieeetr}
\bibliography{ref}




\vspace{-33pt}
\begin{IEEEbiographynophoto}{Jingyou Xie}
received the Master degree from the School of Intelligent Systems Engineering, Sun Yat-Sen University, in 2024. He is mainly focusing on multimodal search and cross-modal retrieval.
\end{IEEEbiographynophoto}

\vspace{-33pt}
\begin{IEEEbiographynophoto}{Jiayi Kuang}
received the BEng degree from the School of Intelligent Systems Engineering, Sun Yat-Sen University, in 2024. She is currently working toward Master’ s degree with the School of Intelligent Systems Engineering, Sun Yat-Sen University and mainly focusing on natural language reasoning, multimodal question answering and multimodal large language models.
\end{IEEEbiographynophoto}

\vspace{-33pt}
\begin{IEEEbiographynophoto}{Zhenzhou Lin}
received the Master degree from the School of Intelligent Systems Engineering, Sun Yat-Sen University, in 2024. He is mainly focusing on natural language processing and knowledge graph sampling.
\end{IEEEbiographynophoto}

\vspace{-33pt}
\begin{IEEEbiographynophoto}{Jiarui Ouyang}
is currently working toward BEng’ s degree with the School of Intelligent Systems Engineering, Sun Yat-Sen University and mainly focusing on large language models.
\end{IEEEbiographynophoto}

\vspace{-33pt}
\begin{IEEEbiographynophoto}{Zishuo Zhao}
received the Master degree from the School of Intelligent Systems Engineering, Sun Yat-Sen University, in 2024. He is mainly focusing on natural language processing and multimodal learning.
\end{IEEEbiographynophoto}

\vspace{-33pt}
\begin{IEEEbiographynophoto}{Ying Shen}
is now an Associate Professor in School of Intelligent Systems Engineering, Sun Yat-Sen University. She received her Ph.D. degree from the University of Paris Ouest Nanterre La Défense (France), specialized in Computer Science. She received her Erasmus Mundus Master degree in Natural Language Processing from France and England. Her research interests include Natural Language Processing and multimodal deep learning.
\end{IEEEbiographynophoto}

\vfill

\end{document}